\documentclass{article}

\PassOptionsToPackage{numbers,sort&compress}{natbib}

\usepackage[preprint]{neurips_2025}

\usepackage[utf8]{inputenc}
\usepackage[T1]{fontenc}

\usepackage{amsfonts}
\usepackage{amsmath}
\usepackage{bm}

\usepackage{algorithm}
\usepackage{algpseudocode}

\usepackage{booktabs}
\usepackage{graphicx}
\usepackage{adjustbox}
\usepackage{makecell}
\usepackage{multirow}
\usepackage{rotating}

\usepackage{microtype}
\usepackage{nicefrac}
\usepackage{pifont}

\usepackage{xcolor}
\usepackage{hyperref}
\usepackage{url}

\usepackage{tikz}
\usetikzlibrary{shapes.geometric, arrows.meta, positioning, calc, tikzmark}

\definecolor{highlightColor}{RGB}{142,22,22}
\definecolor{boxBackgroundColor}{RGB}{248,238,223}

\newcommand{\mytick}{\textcolor{green!70!black}{\ding{52}}} %
\newcommand{\myban}{\textcolor{red!80!black}{\ding{55}}}   %

\title{A Minimum Description Length Approach to Regularization in Neural Networks}
\author{
Matan Abudy\textsuperscript{1}\thanks{Equal contribution.} \quad
Orr Well\textsuperscript{1}\footnotemark[1] \quad
Emmanuel Chemla\textsuperscript{2}\thanks{Equal contribution.} \quad
Roni Katzir\textsuperscript{1}\footnotemark[2] \quad
Nur Lan\textsuperscript{2}\footnotemark[2] \\
\textsuperscript{1}Tel Aviv University \quad \textsuperscript{2}École Normale Supérieure \\
\texttt{matan.abudy@gmail.com, orrwell@mail.tau.ac.il} \\
\texttt{\{emmanuel.chemla,nur.lan\}@ens.psl.eu, rkatzir@tauex.tau.ac.il}
}

\begin{document}
\maketitle

\begin{abstract}
State-of-the-art neural networks can be trained to become remarkable solutions to many problems. But while these architectures \emph{can} express symbolic, perfect solutions, trained models often arrive at approximations instead. We show that the choice of regularization method plays a crucial role: when trained on formal languages with standard regularization ($L_1$, $L_2$, or none), expressive architectures not only fail to converge to correct solutions but are actively pushed away from perfect initializations. In contrast, applying the Minimum Description Length (MDL) principle to balance model complexity with data fit provides a theoretically grounded regularization method. Using MDL, perfect solutions are selected over approximations, independently of the optimization algorithm. We propose that unlike existing regularization techniques, MDL introduces the appropriate inductive bias to effectively counteract overfitting and promote generalization.
\end{abstract}

\section{Introduction}
Modern neural networks can closely approximate solutions to highly complex problems. However, tasks requiring formal reasoning and accurate generalization still show a substantial gap between state-of-the-art models and human intelligence. For example, although models perform well on small-digit arithmetic, they often fail on large-digit numbers in counting, addition, and multiplication \citep{gambardella2024language, yehudai2024can, brown2020language}. They also struggle with multi-step, recursive and compositional reasoning \citep{dziri2023faith, valmeekam2022large, guzman2024testing}, fail tests that break co-occurrence patterns \citep{nezhurina2024alice, wu2024reasoning, bertolazzi2024systematic} and are sensitive to slight changes in task presentation \citep{vafa2025evaluating, mirzadeh2024gsm}. In some cases, even chain-of-thought approaches do not lead to the desired behavior \citep{nezhurina2024alice, dziri2023faith}.

All of these tasks involve underlying rules, syntactic or semantic in nature. Models struggle especially with inductive reasoning, where they are required to infer these rules from observed data \citep{hua2025inductionbench}. This inability to generalize beyond the training distribution could be explained if the models rely on heuristics and approximations rather than learning the actual rules. For some models and tasks, it was shown that this is indeed the case \citep{zhang2022paradox, nikankin2024arithmetic, vafa2025evaluating}. 

Problems may arise from the choice of architecture, objective, or optimization algorithm. Some neural network architectures, including Transformers, are in theory expressive enough to represent correct solutions to a wide range of problems \citep{siegelmann1992computational, weiss2021thinking, elnaggar2023formal, stogin2024provably}. We therefore ask what part of the problem lies in the choice of the objective function itself. As the problem is with generalization, we focus on regularization - the part of the objective meant to counteract overfitting.

We propose Minimum Description Length (MDL) regularization as an inductive bias that can substantially improve performance on tasks that require systematic generalization. Although our work focuses on formal languages, many of the failed tasks mentioned above can be framed in a similar way. Formal languages thus provide a natural benchmark for evaluating generalization in neural networks.

\paragraph{Contributions.} We provide empirical evidence for the adequacy of MDL as a regularization method in neural networks. The current work makes three main contributions: (i) Extending previous work on MDL regularization to context-free grammars beyond counting and to more naturalistic grammars. (ii) Providing a systematic comparison of MDL with $L_1$, $L_2$, and absence of regularization, showing that only MDL consistently preserves or compresses perfect solutions, while other methods push models away from these solutions and degrade their performance. (iii) Establishing the success of MDL across learning settings (architecture search vs.~fixed-architecture training) and optimization algorithms (genetic search vs.~gradient descent). All code and experimental data are publicly available at \url{https://github.com/taucompling/mdl-reg-approach}.

\subsection{Minimum Description Length (MDL)}
The MDL \citep{solomonoff1964formal,rissanen1978modeling} approach to learning avoids overfitting by balancing data fit with model complexity. Maximizing data fit is common to standard training approaches, but MDL imposes an additional requirement of model simplicity, which prevents memorization of accidental patterns. Since identification of meaningful regularities allows compression, data fit is measured by its description length. Similarly, model complexity is measured by its encoding length. 

Formally, we aim to find a model that minimizes the MDL objective $|H| + |D:H|$, where $|H|$ is the encoding length of the hypothesis represented by the model, and $|D:H|$ is the description length of the data under this hypothesis. 

For neural networks, $|D:H|$ under Shannon-Fano coding \citep{shannon1948mathematical} is equivalent to the log surprisal, or cross-entropy (CE) loss: 

\[
- \sum_{i=1}^{n} \sum_{t=1}^{m} \log \left( q(c_{it}) \right)
\]

where $q$ is the model distribution and $c_{it}$ is the target class at time step $t$ in sequence $i$.

Therefore, MDL can be viewed as a form of regularization: minimizing the MDL objective is equivalent to minimizing the standard CE loss, alongside a regularization term that reflects the information content (complexity) of the network.

There are \textit{a priori} reasons to favor MDL over standard regularization. Standard methods constrain weight magnitudes, but even numerically small weights can ``smuggle'' information through high-precision values. Sparsity does not guarantee generalization either - in principle, a network could encode an entire corpus as a bit string into a single weight. MDL offers a principled alternative: unlike standard regularization, it accounts for the full information content of the network and penalizes any form of information smuggling or memorization.

MDL is also cognitively motivated, as human learning behavior aligns with MDL and the related Bayesian framework (assuming a simplicity prior). For example, \citet{orban2008bayesian} show that subjects identify basic elements in images in line with Bayesian predictions. Other experiments show similar generalization patterns in learning new words from very few examples (``fast mapping'') \citep{xu2007word, schmidt2009meaning, tenenbaum2011grow}.

\subsection{Related work}
MDL and related Bayesian methods have been applied to a wide range of linguistic phenomena, where generalization from limited data is crucial \citep{horning1969study, berwick1982locality, stolcke1994bayesian, grunwald1995minimum, de1996unsupervised}. In phonology, the only learners to date that infer adult-like linguistic knowledge from distributional evidence are based on MDL \citep{rasin2016evaluation, rasin2021approaching}.

Using a simplicity criterion for artificial neural networks dates back to \citet{hinton1993keeping, zhang1993evolving, zhang1995balancing} and \citet{schmidhuber1997discovering}. More recently, MDL has been applied to balance memorization and generalization in Hopfield networks \citep{abudy2023minimum}, and to train a ReLU network to classify primes \citep{chatterjee2024neural}.

Most relevant, \citet{lan2022minimum} trained Recurrent Neural Networks (RNNs) with MDL to perfectly learn several formal languages, but failed on the more complex Dyck-2 language due to search limitations. \citet{lan2024bridging} showed that Long Short-Term Memory (LSTM) networks trained on the formal language $a^nb^n$ failed to generalize to the correct grammar with $L_1$ or $L_2$ regularization, even when a perfect solution was reachable via weight tuning; only MDL identified the correct solution as a minimum of the objective function.

\section{Method}

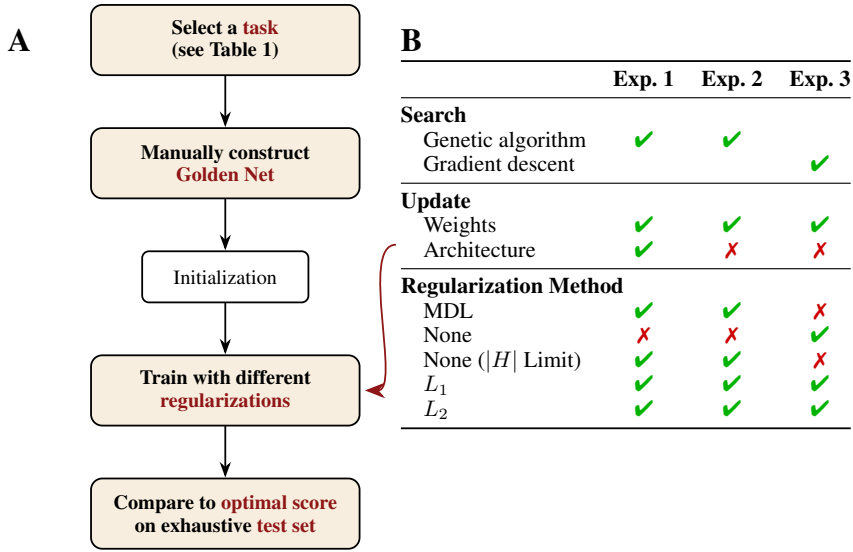
\begin{figure}[htbp]
  \centering
\scalebox{0.85}{
  \begin{tabular}{@{}c@{\hspace{1.8em}}c@{}}
    \begin{tikzpicture}[
        remember picture,
        every node/.style = {align=center},
        mainbox/.style    = {rounded corners=5pt,
                             minimum width=4.2cm,
                             minimum height=1.1cm,
                             fill=boxBackgroundColor,
                             draw=black, thick,
                             font=\bfseries\small
                             },
        initbox/.style    = {rounded corners=3pt,
                             minimum width=2.6cm,
                             minimum height=0.8cm,
                             fill=white,
                             draw=black, thick,
                             font=\small
                             },
        flowarrow/.style  = {-{Stealth[length=7pt, width=5pt]},
                             thick,
                             line cap=round},
        baseline=(labelA.base)
      ]

      \node[anchor=west, font=\bfseries\Large] (labelA) {A};

      \node[mainbox, right=0.8cm of labelA.east] (step1)
            {Select a \textcolor{highlightColor}{task}\\(see Table \ref{tab:tasks})};
      \node[mainbox, below=0.8cm of step1] (step2)
            {Manually construct\\\textcolor{highlightColor}{Golden Net}};
      \node[initbox, below=0.8cm of step2] (init)
            {Initialization};
      \node[mainbox, below=0.8cm of init] (step3)
            {Train with different\\\textcolor{highlightColor}{regularizations}};
      \node[mainbox, below=0.8cm of step3] (step4)
            {Compare to \textcolor{highlightColor}{optimal score}\\on exhaustive
             \textcolor{highlightColor}{test set}};

      \draw[flowarrow] (step1) -- (step2);
      \draw[flowarrow] (step2) -- (init);
      \draw[flowarrow] (init)  -- (step3);
      \draw[flowarrow] (step3) -- (step4);
    \end{tikzpicture}
    &

    \begin{minipage}[t]{0.55\textwidth}
      \makebox[0pt][l]{\textbf{\Large B}}\vspace{0.3em}

      \tikzmarknode{sourceTableInB}{%
      \begin{tabular}{@{}lccc@{}}
        \toprule
        & \textbf{Exp.~1} & \textbf{Exp.~2} & \textbf{Exp.~3}\\
        \midrule
        \multicolumn{4}{@{}l}{\textbf{Search}}\\
        \quad Genetic algorithm   & \mytick & \mytick &         \\
        \quad Gradient descent    &         &         & \mytick \\
        \midrule
        \multicolumn{4}{@{}l}{\textbf{Update}}\\
        \quad Weights             & \mytick & \mytick & \mytick \\
        \quad Architecture        & \mytick & \myban  & \myban  \\
        \midrule
        \multicolumn{4}{@{}l}{\textbf{Regularization Method}}\\
        \quad MDL                 & \mytick & \mytick & \myban  \\
        \quad None                & \myban  & \myban & \mytick \\
        \quad None ($|H|$ Limit)  & \mytick & \mytick & \myban  \\
        \quad $L_1$               & \mytick & \mytick & \mytick \\
        \quad $L_2$               & \mytick & \mytick & \mytick \\
        \bottomrule
      \end{tabular}%
      }
    \end{minipage}
  \end{tabular}

  \begin{tikzpicture}[overlay, remember picture]
    \draw[-{Stealth[length=9pt, width=7pt]}, highlightColor, thick, shorten <=2pt, shorten >=2pt]
      (sourceTableInB.west)             
      to[out=180, in=0, looseness=0.7]  
      (step3.east);                     
  \end{tikzpicture}
}
  \caption{\textbf{Method overview.} We compare the relative benefit of different
           regularization methods with the same pipeline (A) across
           training regimes (B).}
  \label{fig:overview}
\end{figure}

\subsection{Overview}
We define a battery of next-token prediction tasks based on formal languages, each represented by a probabilistic grammar. For each task, we manually construct a ``golden'' RNN that perfectly solves it by matching the true data distribution. Initializing the learning process with this golden network allows us to examine how each regularization method interacts with correct solutions, without requiring the search to discover them from scratch.

We explore three training settings. Experiments 1 and 2 use a genetic algorithm (GA) to evolve networks based on their regularized objective score. Experiment 1 mutates both architecture and parameters, while experiment 2 restricts mutations only to weights and biases. Experiment 3 uses gradient descent (GD) to train the golden network’s weights; MDL is excluded due to its non-differentiability. Hyperparameters are detailed in Appendix~\ref{app:hyperparams}.

Performance is evaluated by comparing CE loss on a test set to the theoretical optimum. Figure~\ref{fig:overview} illustrates the general setup.

\subsection{Architecture}
We use a general formulation of an RNN, which we refer to as a ``free-form RNN''. This is a directed graph consisting of units with biases and weighted connections. Connections may be forward or recurrent, and each unit may have a different activation function.

These networks are at least as expressive as ReLU RNNs (as any ReLU RNN can be represented as a free-form RNN) and are therefore capable of performing tasks that require counting \citep{elnaggar2023formal}, like some of the tasks described below. The existence of golden networks that perfectly solve all of our tasks further supports their expressivity, including capabilities beyond simple counting.

\subsection{Tasks}
Each task is defined by a formal language, with datasets generated from its corresponding probabilistic grammar. Table \ref{tab:tasks} lists representative examples, and full grammar definitions appear in Appendix \ref{app:grammars}.  

The tasks are designed to test particular computational abilities: $a^n b^n$ and Dyck-1 require counting; $a^n b^n c^n$ requires two counters, and Dyck-2 - a stack. Arithmetic involves nested addition formulas over 1’s and requires a stack to track argument positions. Toy-English is a minimal English fragment with one noun phrase, one transitive verb, one intransitive verb, and one sentential complement verb, supporting object relative clauses. Although the last two tasks are meant to represent more naturalistic grammars, belonging to these languages is still a matter of syntax, with no notion of semantic correctness.

\begin{table}
	\caption{Summary of tasks and example strings.}
	\label{tab:tasks}
	\centering
	\begin{tabular}{ll}
		\toprule
		\textbf{Task} & \textbf{Examples}                                \\
		\midrule
		\makecell[tl]{\boldmath$\bm{a^n b^n}$ \\ Equal number of ordered $a$'s and $b$'s} 
        		              & \makecell[tl]{
                                \texttt{ab} \\
                                \texttt{aaabbb}} \\
		\addlinespace
		\addlinespace
		\makecell[tl]{\boldmath$\bm{a^n b^n c^n}$ \\ Equal number of ordered $a$'s, $b$'s and $c$'s} 
        		              & \makecell[tl]{
                                \texttt{abc} \\
                                \texttt{aaabbbccc}} \\
		\addlinespace
		\addlinespace
		\makecell[tl]{\textbf{Dyck-1} \\ Well-matched parentheses} 
		              & \makecell[tl]{\texttt{()()}\\
                            \texttt{(((()))(()))}} \\
		\addlinespace
		\addlinespace
		\makecell[tl]{\textbf{Dyck-2} \\ Well-matched parentheses and brackets} 
		              & \makecell[tl]{\texttt{()[]}\\
                        \texttt{[()](([()]))}} \\
		\addlinespace
		\addlinespace
		\makecell[tl]{\textbf{Arithmetic Syntax} \\ Nested addition formulas} 
		              & \makecell[tl]{\texttt{((1+1)+(1+1))}\\
                      \texttt{(((1+((1+1)+1))+1)+1)}} \\
		\addlinespace
		\addlinespace
		\makecell[tl]{\textbf{Toy-English} \\ Fragment of English with relative clauses} 
		              & \makecell[tl]{\texttt{dogs think that dogs sleep} \\ \texttt{dogs think that dogs chase dogs} 
                      \\ \texttt{dogs chase dogs dogs chase}} \\
		\bottomrule
	\end{tabular}
\end{table}

\subsection{Golden networks}
We adopt a strong notion of correctness: a golden network not only assigns the highest probability to the correct next symbol at each timestep; its entire output probability distribution matches the true grammar distribution.

For $a^n b^n$, $a^n b^n c^n$ and Dyck-2, we use golden networks previously discovered by \citet{lan2022minimum}. The golden network for Dyck-1 was found in a GA run under MDL. For Arithmetic Syntax and Toy-English, we constructed golden networks manually and empirically verified their perfect performance. Network diagrams are available in Appendix \ref{app:nets}.

\subsection{Regularization methods}
\paragraph{MDL.}
Following \citet{lan2022minimum}, we calculate the regularization term $|H|$ by encoding the structure of the network: unit count, each unit's type, activation and bias, and the connections (recurrent vs.~forward). 

Weights and biases are stored as signed fractions, with bit strings formed by concatenating the sign bit, numerator and denominator (each encoded using the prefix-free scheme from \citet{li2008introduction}). This encoding reflects an intuitive notion of simplicity that goes beyond absolute size: a weight is simple if it has a concise fractional form. For example, $0.1$ encodes as $1/10$ (simple), while the smaller $0.02234$ expands to $1117/50{,}000$ (complex). Thus, $1/10$ receives a shorter code. This also illustrates that a naive binary encoding fails to capture our intuitions - $1/10$ is simple despite lacking an exact base-2 floating-point representation.

\paragraph{$\bm{L_1}$.} 
A regularization method that encourages sparsity by penalizing absolute weight values:
\[
\lambda \sum_{i=1}^{d} |w_i|
\]
where \( \lambda \) is the regularization coefficient.

\paragraph{$\bm{L_2}$.} 
A regularization method aimed at keeping weights small by penalizing their squared values:
\[
\lambda \sum_{i=1}^{d} w_i^2
\]

\paragraph{No regularization (with a limit on $\bm{|H|}$).}
Minimizing CE loss alone leads to overfitting, as networks can grow indefinitely and memorize the training set. In GA experiments, ``exploding'' networks can cause overload and slowness. To prevent this, we impose a complexity ceiling - three times the golden network’s $|H|$ — which charitably introduces an additional level of regularization.

\subsection{Evaluation}
Accuracy can be misleading: A model can reach 100\% next-token accuracy via \texttt{argmax} prediction, while misrepresenting the true distribution. And when the underlying grammar is non-deterministic, even a model that matches the true distribution may assign the highest probability to an incorrect symbol (for further discussion see \citet{lan2023benchmarking}). Instead of accuracy, we evaluate models using their test set $|D:H|$, equivalent to CE loss. To avoid infinite scores (due to $\log 0$), we smooth the network output distribution by adding $10^{-10}$ to zero probabilities. 

Sampling test strings from the grammar, as done for training, may miss low-probability strings with high surprisal; therefore, we construct exhaustive test sets that contain all strings up to a length threshold, weighted by their true probabilities. These test sets approximate the full grammar distribution; while low training $|D:H|$ may indicate overfitting, low test $|D:H|$ should reflect generalization.

However, as the test set is still a finite approximation, overfitting it remains possible. Thus, we evaluate models not by the absolute value of their test $|D:H|$ score, but by its proximity to the analytically computed optimal score, derived by summing log probabilities under the true distribution. Although a test score below the optimum does not necessarily reflect a better solution, one above the optimum clearly reflects a flawed solution.

\section{Experiments and results}
\subsection{Experiment 1: Genetic architecture search}
Training with GD is unsuitable for MDL, as the $|H|$ term is non-differentiable. Following \citet{lan2022minimum}, we optimize networks using a genetic algorithm (GA) \citep{holland1992adaptation}. We use the Island Model \citep{gordon1993serial, adamidis1994review, cantu1998survey}, where the population is divided into equal ``islands'' that evolve independently with occasional migration. Within each island, networks are mutated and selected based on their regularized objective score. Pseudo-code appears in Appendix~\ref{app:gen-alg}.

The GA mutates networks by adding or removing units or connections, modifying weights and biases, and changing activation functions. Allowed activations include linear, ReLU, $\tanh$, and task-specific functions used by the golden network, enabling the search to leverage the same activations.

\begin{table}
    \caption{\textbf{GA architecture search results.} For each task and regularization method, we evaluate the GA‑selected network on training samples and an exhaustive test set, reporting the relative gap $\Delta (\%) = \frac{|D:H| - \text{Optimal}}{\text{Optimal}} \times 100$. * marks originally infinite scores due to zero‑probability errors. MDL yields the smallest test gaps and typically preserves or even compresses the golden network.}
    \label{tab:ga-search-results}
	\centering
    \begin{adjustbox}{max width=\textwidth}
\begin{tabular}{llccccc}
\toprule
Task & Regularizer & $|H|$ & Train $|D:H|$ & Test $|D:H|$ & $\Delta^{Train}_{Optim}$ (\%) & $\Delta^{Test}_{Optim}$ (\%) \\
\midrule
a$^n$b$^n$ & (Golden) & (139) & (1531.77) & (2.94) & (0.0) & (0.0) \\
& MDL (|H|) & 139 & 1529.39 & 2.94 & \textbf{-0.2} & \textbf{0.1} \\
& $L_1$ & 460 & 1498.12 & 2.94$^*$ & -2.2 & 0.2 \\
& $L_2$ & 600 & 1514.66 & 3.27$^*$ & -1.1 & 11.4 \\
& None (Lim. $|H|$) & 430 & 1488.17 & 3.32$^*$ & -2.8 & 13.0 \\
\midrule
a$^n$b$^n$c$^n$ & (Golden) & (241) & (1483.91) & (2.94) & (0.0) & (0.0) \\
& MDL (|H|) & 241 & 1483.91 & 2.94 & \textbf{0.0} & \textbf{0.0} \\
& $L_1$ & 366 & 1482.88 & 2.94 & -0.1 & 0.1 \\
& $L_2$ & 602 & 1476.71 & 3.03$^*$ & -0.5 & 3.2 \\
& None (Lim. $|H|$) & 736 & 1460.67 & 2.96 & -1.6 & 0.6 \\
\midrule
Dyck-1 & (Golden) & (113) & (1544.48) & (1.77) & (-0.0) & (0.0) \\
& MDL (|H|) & 113 & 1544.22 & 1.77 & \textbf{-0.0} & \textbf{0.1} \\
& $L_1$ & 753 & 1488.17 & 1.90$^*$ & -3.6 & 7.5 \\
& $L_2$ & 1140 & 1479.37 & 1.96$^*$ & -4.2 & 10.7 \\
& None (Lim. $|H|$) & 353 & 1505.61 & 2.03$^*$ & -2.5 & 15.1 \\
\midrule
Dyck-2 & (Golden) & (579) & (2121.53) & (2.32) & (-0.0) & (0.0) \\
& MDL (|H|) & 327 & 2130.28 & 2.37 & \textbf{0.4} & \textbf{2.0} \\
& $L_1$ & 1629 & 2009.61 & 2.70$^*$ & -5.3 & 16.1 \\
& $L_2$ & 2248 & 1996.82 & 2.82$^*$ & -5.9 & 21.3 \\
& None (Lim. $|H|$) & 1757 & 1992.89 & 2.61$^*$ & -6.1 & 12.5 \\
\midrule
Arithmetic & (Golden) & (967) & (2581.29) & (3.96) & (0.0) & (0.1) \\
& MDL (|H|) & 431 & 2581.07 & 3.94 & \textbf{0.0} & \textbf{-0.4} \\
& $L_1$ & 1268 & 2540.22 & 3.99$^*$ & -1.6 & 1.0 \\
& $L_2$ & 1200 & 2560.62 & 4.01$^*$ & -0.8 & 1.4 \\
& None (Lim. $|H|$) & 3004 & 2477.42 & 4.23$^*$ & -4.0 & 7.0 \\
\midrule
Toy English & (Golden) & (870) & (2232.74) & (4.49) & (0.0) & (0.0) \\
& MDL (|H|) & 414 & 2231.96 & 4.57$^*$ & \textbf{-0.0} & \textbf{2.0} \\
& $L_1$ & 1215 & 2209.25 & 5.05$^*$ & -1.0 & 12.6 \\
& $L_2$ & 1330 & 2202.41 & 5.43$^*$ & -1.3 & 21.0 \\
& None (Lim. $|H|$) & 2625 & 2142.40 & 6.25$^*$ & -4.0 & 39.4 \\
\bottomrule
\end{tabular}
\end{adjustbox}
\end{table}

\begin{figure}[ht]
    \centering
    \includegraphics[width=\linewidth]{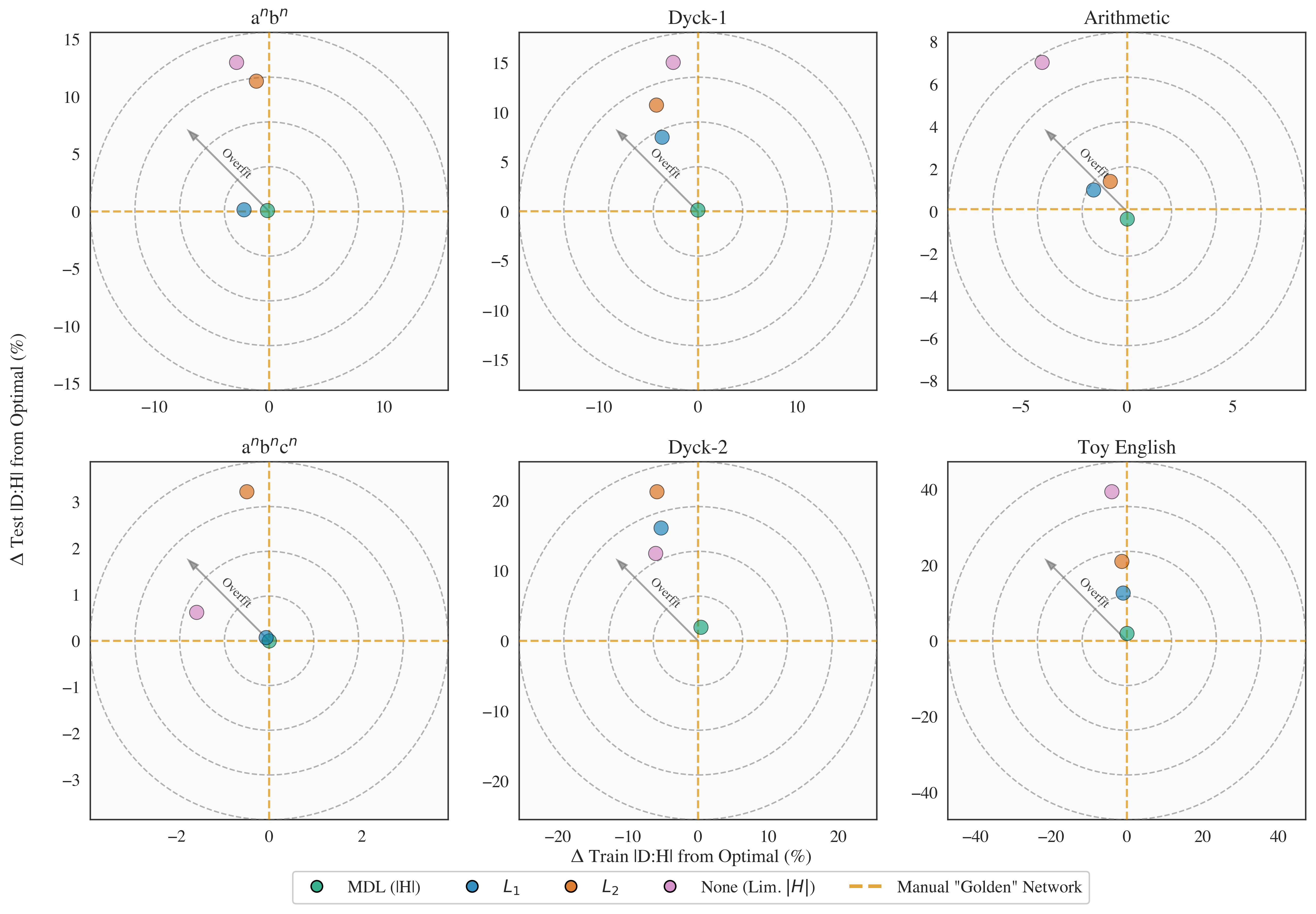}
    \caption{Relative deviation ($\Delta\%$) from optimal $|D:H|$ for each final network in Experiment 1, on train (x-axis) and test (y-axis), grouped by task. Proximity to center indicates better approximation of the analytical optimum.}
    \label{fig:exp-1-target-plot}
\end{figure}

Table~\ref{tab:ga-search-results} reports final scores for network complexity ($|H|$) and data fit ($|D:H|$) with regard to the training and test sets, as well as the relative deviation from the optimal score $\Delta(\%) = \frac{|D:H| - \text{Optimal}}{\text{Optimal}} \times 100$. A positive delta indicates underfitting, while a negative value indicates overfitting. Asterisks mark scores that were originally infinite due to a zero probability assigned to the correct next symbol, corrected via smoothing. Figure~\ref{fig:exp-1-target-plot} visualizes these deviations.

MDL-selected networks yield the lowest and closest to optimal test $|D:H|$ scores across all tasks. In addition, MDL tends to preserve the golden network during search, unlike the other regularizers. However, when golden networks are constructed manually they can be much larger than necessary; in those cases, MDL compresses them significantly while retaining near-optimal $|D:H|$, producing more compact models.

\subsection{Experiment 2: GA in weight-training setting}
The genetic algorithm can modify the network architecture during optimization, making it more akin to an architecture search procedure than to standard training. One might argue that since $L_1$ and $L_2$ are typically applied during training with a fixed architecture, comparisons to MDL should also be made in that setting.

We believe that the comparison in experiment 1 is justified: in practice, regularization also guides architecture search, as the process usually also involves weight training. However, we can also compare MDL to standard regularization in a pure training setting. To do that, we initialize the GA population solely with instances of the golden network and restrict mutations to weights and biases, keeping the architecture fixed.

Under these constraints, MDL again outperforms $L_1$ and $L_2$, confirming that it is the appropriate regularizer even when the architecture is fixed.

\begin{table}
    \caption{\textbf{GA weight-training results.} For each task and regularization method, we evaluate the network selected by GA when only mutations of weights and biases are allowed. We measure performance on training samples and on an exhaustive test set, reporting the relative gap $\Delta (\%) = \frac{|D:H| - \text{Optimal}}{\text{Optimal}} \times 100$; * marks originally infinite scores due to zero‑probability errors. Once again, MDL yields the smallest gaps from the optimum and tends to preserve or simplify the golden network.}
    \label{tab:ga-weight-results}
	\centering
    \begin{adjustbox}{max width=\textwidth}
\begin{tabular}{llccccc}
\toprule
Task & Regularizer & $|H|$ & Train $|D:H|$ & Test $|D:H|$ & $\Delta^{Train}_{Optim}$ (\%) & $\Delta^{Test}_{Optim}$ (\%) \\
\midrule
a$^n$b$^n$ & (Golden) & (139) & (1531.77) & (2.94) & (0.0) & (0.0) \\
& MDL (|H|) & 139 & 1529.39 & 2.94 & \textbf{-0.2} & \textbf{0.1} \\
& $L_1$ & 217 & 1528.92 & 2.94 & -0.2 & 0.2 \\
& $L_2$ & 268 & 1528.96 & 2.94 & -0.2 & 0.1 \\
& None (Lim. $|H|$) & 385 & 1524.71 & 2.94 & -0.5 & 0.2 \\
\midrule
a$^n$b$^n$c$^n$ & (Golden) & (241) & (1483.91) & (2.94) & (0.0) & (0.0) \\
& MDL (|H|) & 239 & 1483.91 & 2.94 & \textbf{0.0} & \textbf{0.0} \\
& $L_1$ & 362 & 1482.88 & 2.94 & -0.1 & 0.1 \\
& $L_2$ & 445 & 1482.94 & 2.94 & -0.1 & 0.0 \\
& None (Lim. $|H|$) & 733 & 1481.57 & 2.94 & -0.2 & 0.1 \\
\midrule
Dyck-1 & (Golden) & (113) & (1544.48) & (1.77) & (-0.0) & (0.0) \\
& MDL (|H|) & 113 & 1544.22 & 1.77 & \textbf{-0.0} & \textbf{0.1} \\
& $L_1$ & 259 & 1537.35 & 1.79 & -0.5 & 1.1 \\
& $L_2$ & 408 & 1537.28 & 1.79 & -0.5 & 1.2 \\
& None (Lim. $|H|$) & 355 & 1534.32 & 1.79 & -0.7 & 1.3 \\
\midrule
Dyck-2 & (Golden) & (579) & (2121.53) & (2.32) & (-0.0) & (0.0) \\
& MDL (|H|) & 490 & 2121.53 & 2.32 & \textbf{-0.0} & \textbf{0.0} \\
& $L_1$ & 1202 & 2064.86 & 2.39$^*$ & -2.7 & 2.8 \\
& $L_2$ & 1426 & 2055.48 & 2.58$^*$ & -3.1 & 11.2 \\
& None (Lim. $|H|$) & 1731 & 2031.26 & 2.78$^*$ & -4.3 & 19.5 \\
\midrule
Arithmetic & (Golden) & (967) & (2581.29) & (3.96) & (0.0) & (0.1) \\
& MDL (|H|) & 635 & 2581.29 & 3.96 & \textbf{0.0} & \textbf{0.1} \\
& $L_1$ & 1420 & 2555.63 & 3.97$^*$ & -1.0 & 0.4 \\
& $L_2$ & 1553 & 2548.16 & 4.15$^*$ & -1.3 & 5.1 \\
& None (Lim. $|H|$) & 2890 & 2471.51 & 4.79$^*$ & -4.2 & 21.3 \\
\midrule
Toy English & (Golden) & (870) & (2232.74) & (4.49) & (0.0) & (0.0) \\
& MDL (|H|) & 552 & 2237.59 & 4.52 & \textbf{0.2} & \textbf{0.8} \\
& $L_1$ & 1491 & 2186.61 & 5.05$^*$ & -2.1 & 12.7 \\
& $L_2$ & 1704 & 2182.59 & 6.25$^*$ & -2.2 & 39.3 \\
& None (Lim. $|H|$) & 2610 & 2158.71 & 6.21$^*$ & -3.3 & 38.4 \\
\bottomrule
\end{tabular}
    \end{adjustbox}
\end{table}

\subsection{Experiment 3: Gradient descent}
One might object that the optimization algorithm plays an important role in promoting generalization, and propose that unlike GA, GD allows standard regularization to avoid overfitting. While the dynamics of GD are not fully understood, we argue that it cannot fix a flawed objective. This has already been shown for LSTMs on $a^nb^n$ \citep{lan2024bridging}; we extend the analysis to new tasks and free-form RNNs.

Unlike GA, GD cannot handle non-differentiable activations. For tasks that require such activations, GD is therefore inapplicable; but for $a^nb^n$, $a^nb^nc^n$, and Dyck-1, we found golden networks that only use differentiable activations. We train them via backpropagation, starting from the golden weights. Since MDL cannot be optimized with GD, we compare training with $L_1$, $L_2$, or no regularization.

Table~\ref{tab:backprop_results} reports the results. Since the architecture is fixed under GD, we approximate $|H|$ by encoding the weights. We observe that standard regularization consistently drifts away from the golden solution and degrades its performance on the test set.

\begin{table}
    \caption{\textbf{Gradient descent results.} Starting from the golden weights, we train using each differentiable regularizer and report the relative gap $\Delta (\%) = \frac{|D:H| - \text{Optimal}}{\text{Optimal}} \times 100$. All objectives drift away from the golden solution and degrade its performance.}
    \label{tab:backprop_results}
    \centering
    \begin{adjustbox}{max width=\textwidth}
\begin{tabular}{llccccc}
\toprule
Task & Regularizer & $\approx |H|$ & Train $|D:H|$ & Test $|D:H|$ & $\Delta^{Train}_{Optim}$ (\%) & $\Delta^{Test}_{Optim}$ (\%) \\
\midrule
a$^n$b$^n$ & (Golden) & (274) & (1532.33) & (2.94) & (0.0) & (0.1) \\
& None & 5524 & 1529.48 & 2.94 & -0.1 & 0.2 \\
& $L_1$ & 5854 & 1543.98 & 2.95 & 0.8 & 0.5 \\
& $L_2$ & 5610 & 1531.69 & 2.95 & -0.0 & 0.3 \\
\midrule
a$^n$b$^n$c$^n$ & (Golden) & (599) & (1484.15) & (2.94) & (0.0) & (0.0) \\
& None & 20043 & 1482.91 & 2.94 & -0.1 & 0.1 \\
& $L_1$ & 20541 & 1491.15 & 2.96 & 0.5 & 0.9 \\
& $L_2$ & 19909 & 1487.84 & 3.00 & 0.3 & 2.1 \\
\midrule
Dyck-1 & (Golden) & (148) & (1544.28) & (1.78) & (-0.0) & (0.8) \\
& None & 4876 & 1544.03 & 1.78 & -0.0 & 1.0 \\
& $L_1$ & 5138 & 1561.23 & 1.86 & 1.1 & 5.0 \\
& $L_2$ & 4874 & 1562.96 & 1.84 & 1.2 & 4.4 \\
\bottomrule
\end{tabular}
\end{adjustbox}
\end{table}

\section{Discussion}
Our golden networks demonstrate that free‑form RNNs can express exact solutions for every task we studied. Ideally, a golden network should correspond to a (at least local) minimum: for an appropriate objective, departing from a perfect solution is acceptable only if it leads to an equivalent or better solution, so training initialized with a golden network should converge to a network that is at least as good. Across all of our experiments, only the MDL-based objective behaved in this appropriate manner.

For an infinite language, CE loss only approximates true error asymptotically; with finite data, minimizing CE alone leads to overfitting. We show that standard regularization does not prevent this: for every model we trained with $L_1$ or $L_2$, test‑set CE increased after training.

Because they focus solely on weight magnitudes, $L_1$ and $L_2$ ignore other important aspects of network complexity, such as the number of units, activations and parameter information content. Even when only weights and biases are manipulated during the search, MDL mitigates overfitting better than standard regularization: minimizing the MDL term $|H|$ prioritizes weights that are simple to describe and penalizes any form of memorization through small but high-precision weights.

Gradient descent (GD) is sometimes claimed to induce a bias toward generalization \citep{cohen2022implicit, mingard2025deep}. We have shown that even with GD, standard regularization does not reach perfect solutions; in fact, these solutions are not even local minima of the regularized objectives. Success with GD may simply result from imperfect optimization: since the CE loss of neural networks is not convex, GD does not fully minimize it - otherwise, we would have gotten a fully overfitting model. But taking the wrong objective and only partially minimizing it cannot make it right; and it is not surprising that better results are obtained when the right regularization term is chosen from the start.

Even LLMs face the need to generalize from a small amount of data in some practical scenarios, e.g., in-context learning \citep{brown2020language}; formal languages provide a natural benchmark because they require exact generalization. The fact that models struggle, despite expressive architecture and effective optimization, indicates that the problem lies in the objective. The same problem is likely to show up in more complex tasks as well: if a model cannot acquire a simple formal grammar that we know it can express, then it may approximate natural language extremely well, but it is unlikely that it will fully model its underlying structure. 

Indeed, we propose that the choice of regularization also underlies the current model failures mentioned in the introduction. Prior work \citep{lan2024bridging} demonstrated that MDL outperforms $L_1$ and $L_2$ in LSTMs. We extended these findings to free-form RNNs. Although it has not yet been tested empirically, MDL-based regularization could yield the same benefits for Transformers as well.

Some may object that approximations have proven good enough in real-world tasks such as natural language processing. While approximations are inevitable in noisy or ambiguous settings, an approximation that follows the general rule despite some noise is not the same as one that embodies a misunderstanding of the rule itself. Some MDL-selected networks approximate operations (e.g. true zero probabilities are approximated using Softmax) but also preserve the underlying rule structure by effectively implementing mechanisms like counters and stacks.

These results also relate to a deeper cognitive issue: good models should support human-like inductive inference. Although humans also make mistakes, their errors often stem from processing limitations rather than lack of knowledge. Humans use heuristics and approximations in their habitual processing (sometimes referred to as \textit{system 1} \citep{kahneman2011thinking}), enabling them to handle complex tasks despite limited attention and working memory. But they also have mechanisms for controlled and accurate processing (\textit{system 2}), which allow them to generalize well in unfamiliar settings. When these mechanisms are put into action, some errors may vanish. Notably, current models struggle with exactly the kind of tasks that require system 2 processing \citep{goyal2022inductive}, but unlike humans, it is not clear that the models even possess the relevant knowledge, independent of real-time performance. We propose that MDL-based regularization may enable artificial neural networks to also succeed in such cases.

MDL offers a principled framework for designing learning objectives that really aim for perfect generalization. By promoting generalization, regularization based on MDL may allow models to learn from smaller amounts of data. And since it selects models that are smaller, the resulting models are potentially simpler to analyze, which can improve interpretability.

\section{Limitations}
Currently, the main limitation is that the MDL objective is not differentiable, making it better suited to evolutionary search methods that have not yet matched the hardware and software optimizations available for GD. The promise of stronger generalization thus motivates two complementary paths: (i) hardware and software advances that enable non‑differentiable optimization to run at better speeds and efficiency, and (ii) differentiable approximations or surrogate losses that preserve MDL’s bias toward simplicity and generalization.

Additionally, the tasks we evaluated are relatively small in scale. This allowed us to ensure that they were perfectly understood, but future work should explore the benefits of MDL-based regularization when extended to larger-scale settings, additional architectures, and broader benchmarks.

\section{Acknowledgments}
This project was provided with computer and storage resources by GENCI at IDRIS thanks to the grant AD011013783R2 on the supercomputer Jean Zay’s V100 and CSL partitions. RK has been supported by ISF grant \#1083/23.

\bibliographystyle{plainnat}
\bibliography{references}

\appendix
\section{Task-specific probabilistic grammars}
\label{app:grammars}

We generate training strings using probabilistic grammars, most of them context-free, and evaluate generalization using exhaustive test sets that enumerate all valid strings up to a specified length or count. Below we describe the grammar and test set construction for each task.

\subsection*{\bm{$a^n b^n$}, \bm{$a^n b^n c^n$}}
We sample $n \sim \text{Geometric}(p = 0.3)$ and produce either $a^n b^n$ or $a^n b^n c^n$. The test set exhaustively includes strings for all values from $n=1$ up to the largest $n$ in the training set, plus 1000 additional values. For example, if $n_\text{max} = 20$ in training, test includes all $n = 1,\dots,1020$.

\subsection*{Dyck-1, Dyck-2}
Training examples are valid Dyck-1 or Dyck-2 strings sampled from the grammars in Tables~\ref{app:dyck-1-grammar} and~\ref{app:dyck-2-grammar} respectively. To avoid excessive runtime and memory load, the maximal possible training string length is 200. The exhaustive test set includes all Dyck-1 or Dyck-2 strings, depending on the task, of length up to 10.

\begin{table}[H]
\caption{Dyck-1 grammar for generating strings}
\label{app:dyck-1-grammar}
\centering
\begin{tabular}{ll}
\toprule
Production Rule & Probability \\
\midrule
$S \rightarrow [\, S\, ]\, S$ & 0.33333 \\
$S \rightarrow \varepsilon$ & 0.66667 \\
\bottomrule
\end{tabular}
\end{table}

\begin{table}[H]
\caption{Dyck-2 grammar for generating strings}
\label{app:dyck-2-grammar}
\centering
\begin{tabular}{ll}
\toprule
Production Rule & Probability \\
\midrule
$S \rightarrow [\, S\, ]\, S$ & 0.166665 \\
$S \rightarrow (\, S\, )\, S$ & 0.166665 \\
$S \rightarrow \varepsilon$ & 0.66667 \\
\bottomrule
\end{tabular}
\end{table}

\subsection*{Arithmetic Syntax}
Training strings are sampled from the grammar in Table \ref{app:arithmetic-grammar}. The test set enumerates all syntactically valid expressions of length up to 40.

\begin{table}[H]
\caption{Arithmetic Syntax grammar for generating strings}
\label{app:arithmetic-grammar}
\centering
\begin{tabular}{ll}
\toprule
Production Rule & Probability \\
\midrule
$S \rightarrow (\text{E} + \text{E})$ & 1.0 \\
$E \rightarrow 1$ & 0.67 \\
$E \rightarrow (\text{E} + \text{E})$ & 0.33 \\
\bottomrule
\end{tabular}
\end{table}

\subsection*{Toy-English}
Training data consists of strings up to length 200. The test set includes all grammatical strings up to length 20.

\begin{table}[H]
\caption{Toy-English grammar for generating strings}
\label{app:toy-english-grammar}
\centering
\begin{tabular}{ll}
\toprule
Production Rule & Probability \\
\midrule
$S \rightarrow \text{NP VP}$ & 1.0 \\
$\text{NP} \rightarrow \text{N}$ & 0.75 \\
$\text{NP} \rightarrow \text{N RC}$ & 0.25 \\
$\text{RC} \rightarrow \text{NP Vtr}$ & 1.0 \\
$\text{VP} \rightarrow \text{Vtr NP}$ & 0.34 \\
$\text{VP} \rightarrow \text{Vi}$ & 0.33 \\
$\text{VP} \rightarrow \text{Vcp S}$ & 0.33 \\
$\text{N} \rightarrow \text{dogs}$ & 1.0 \\
$\text{Vtr} \rightarrow \text{chase}$ & 1.0 \\
$\text{Vi} \rightarrow \text{sleep}$ & 1.0 \\
$\text{Vcp} \rightarrow \text{think that}$ & 1.0 \\
\bottomrule
\end{tabular}
\end{table}

\section{Golden networks}
\label{app:nets}

\begin{figure}[H]
\centering
\includegraphics[width=0.6\linewidth]{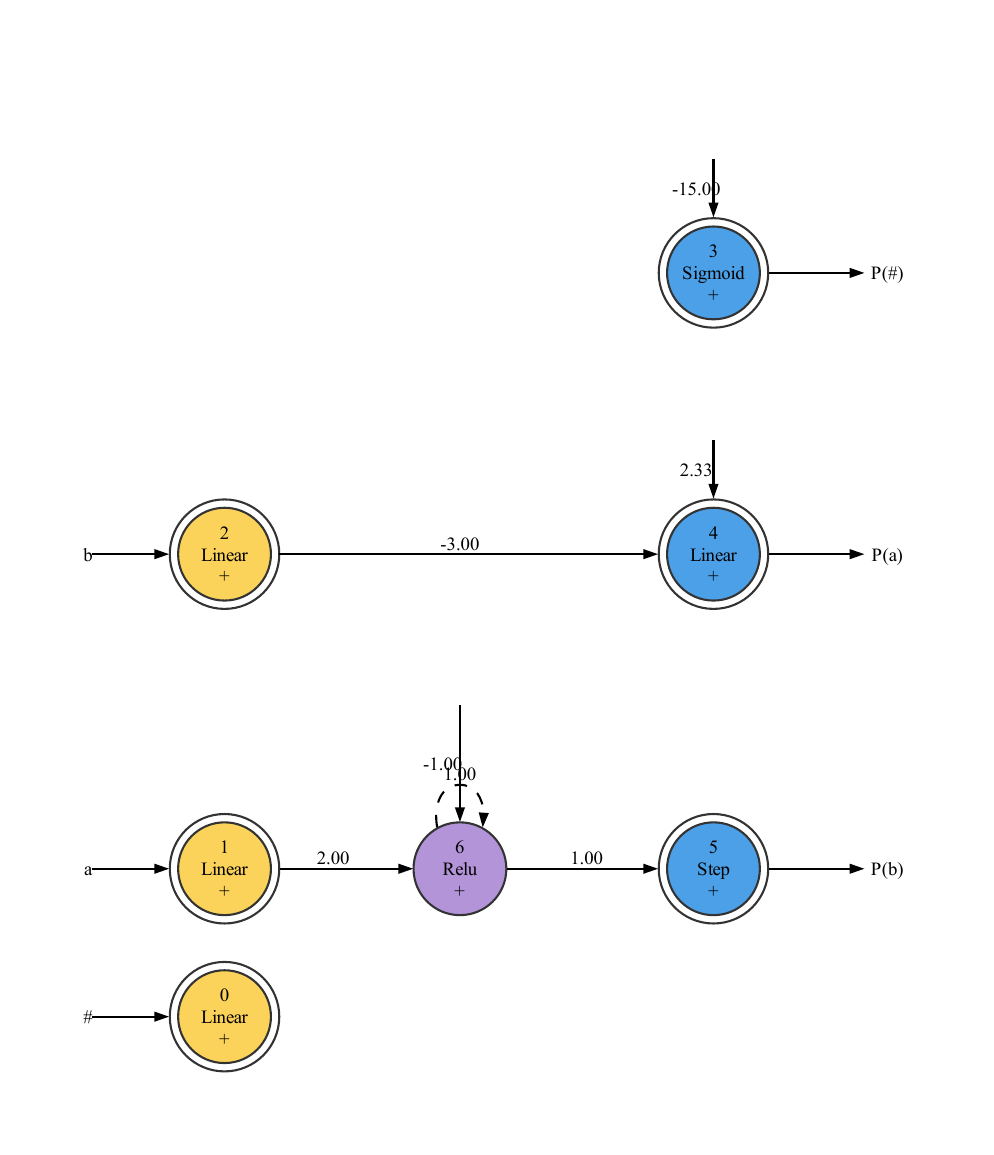}
\caption{Golden network for the $a^n b^n$ task.}
\end{figure}

\begin{figure}[H]
\centering
\includegraphics[width=0.6\linewidth]{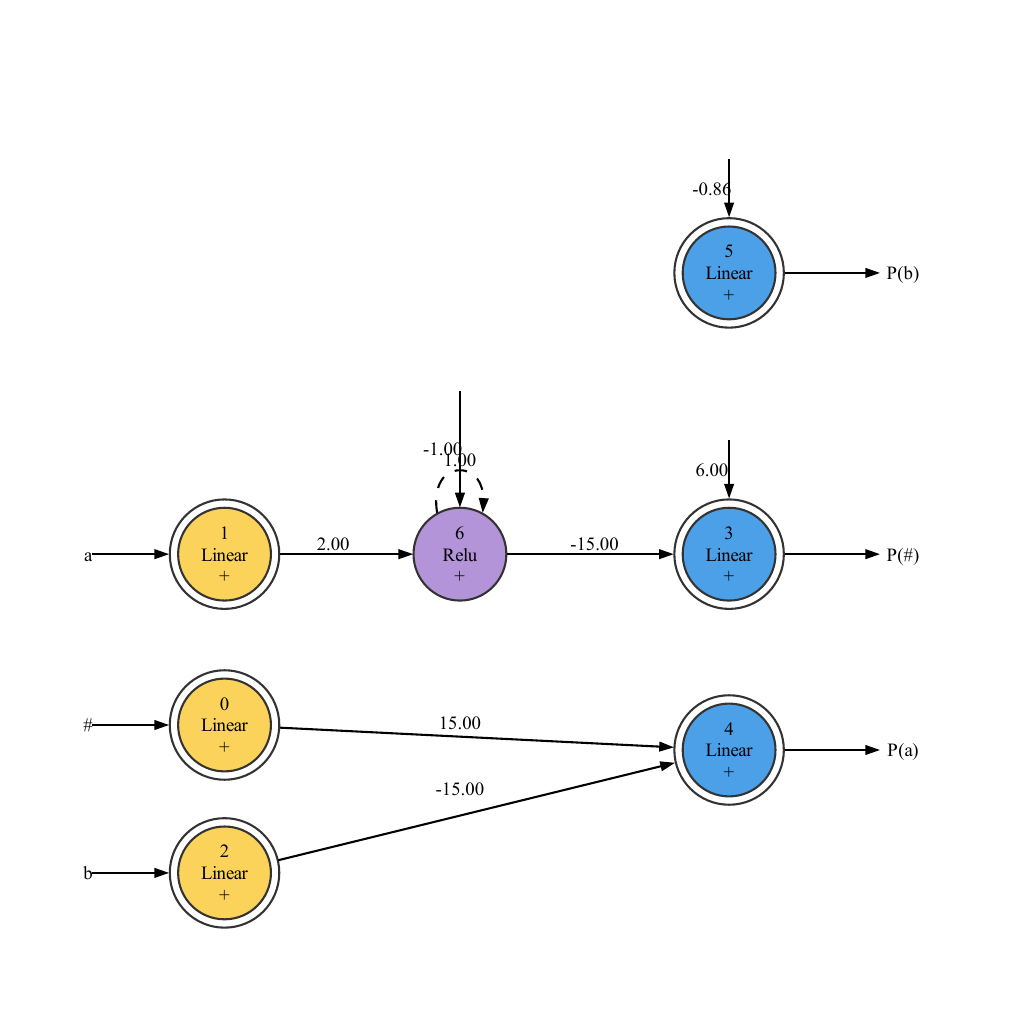}
\caption{Differentiable golden network for the $a^n b^n$ task, used in Experiment 3.}
\end{figure}

\begin{figure}[H]
\centering
\includegraphics[width=0.8\linewidth]{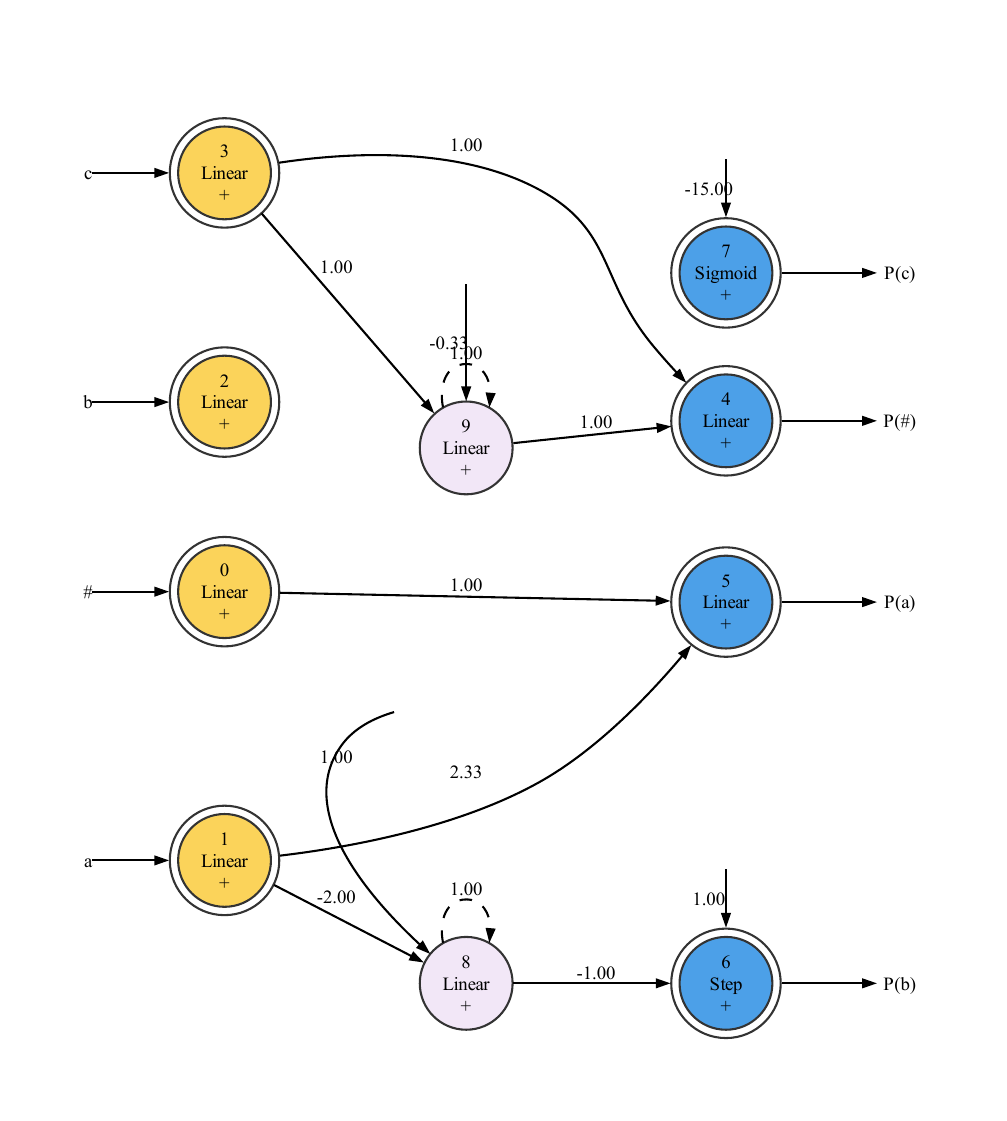}
\caption{Golden network for the $a^n b^n c^n$ task.}
\end{figure}

\begin{figure}[H]
\centering
\includegraphics[width=0.8\linewidth]{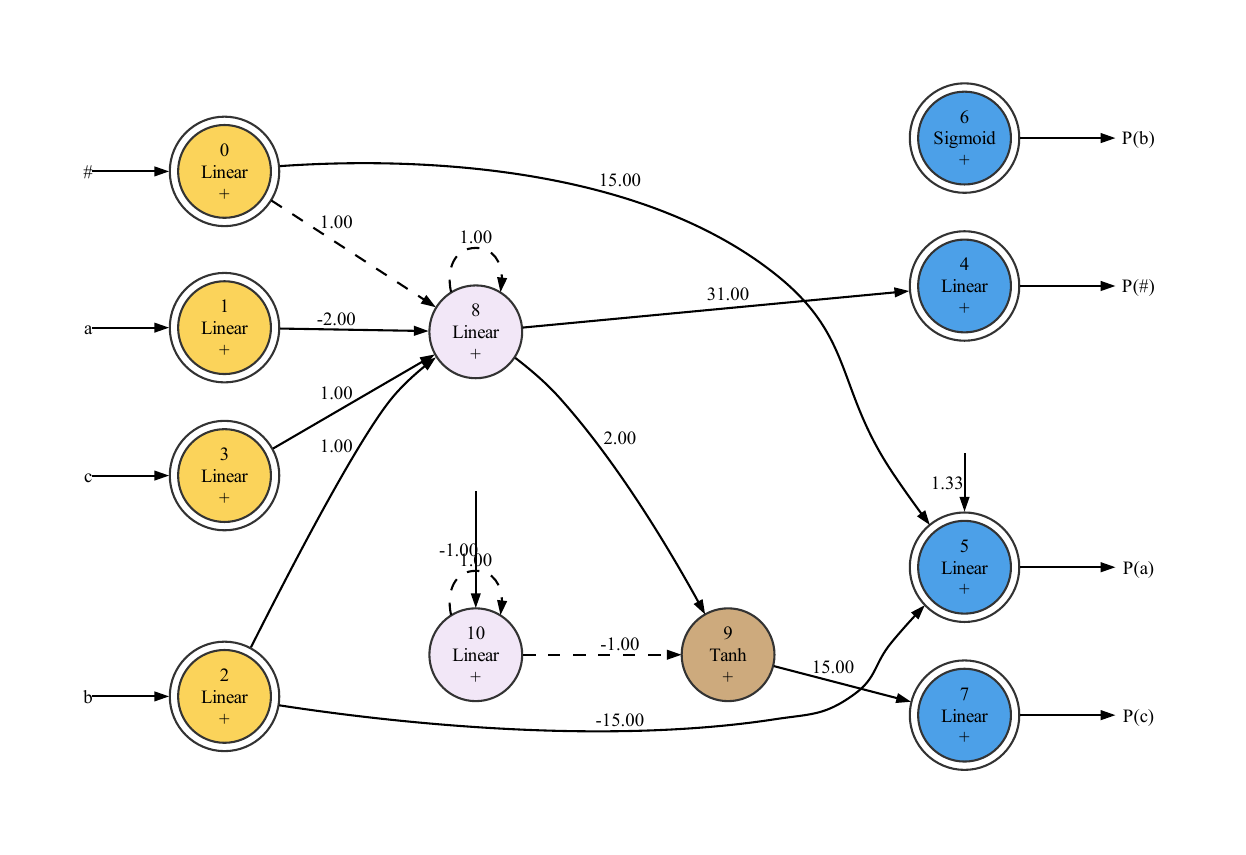}
\caption{Differentiable golden network for the $a^n b^n c^n$ task, used in Experiment 3.}
\end{figure}

\begin{figure}[H]
\centering
\includegraphics[width=0.8\linewidth]{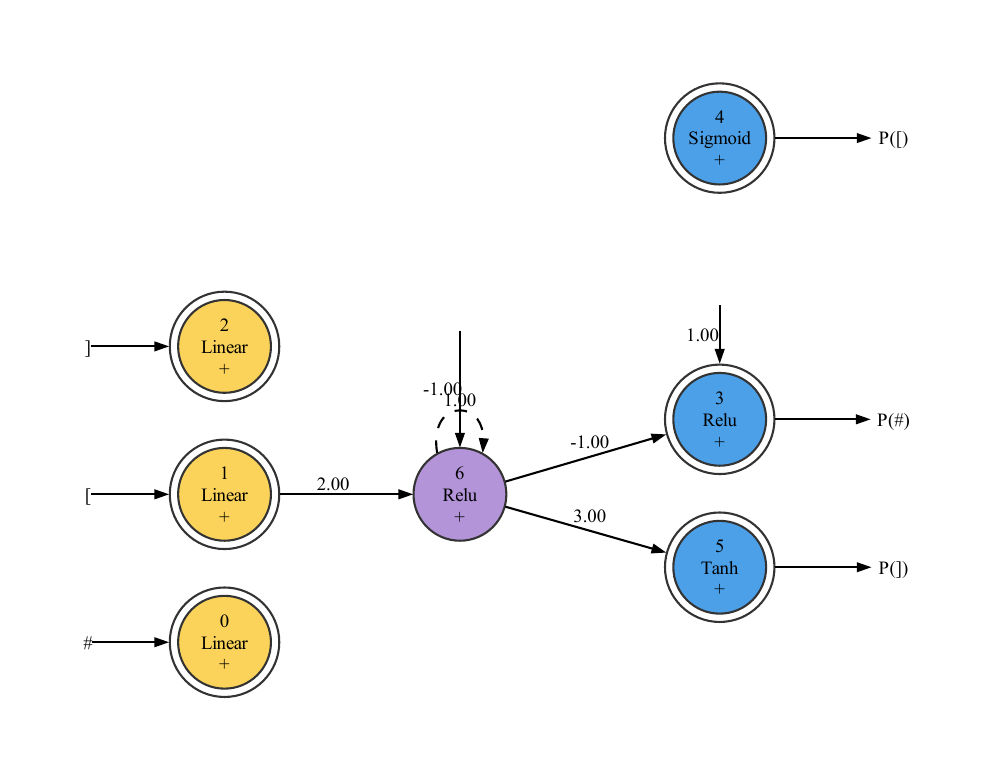}
\caption{Golden network for the Dyck-1 task.}
\end{figure}

\begin{figure}[H]
\centering
\includegraphics[width=0.8\linewidth]{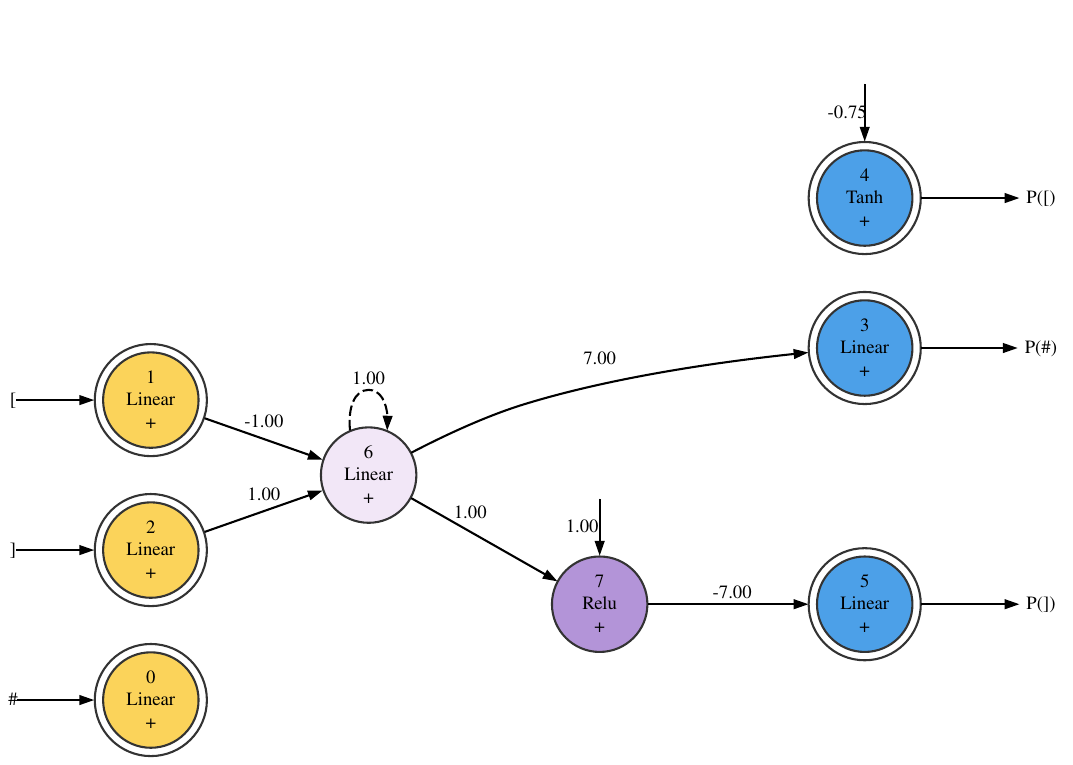}
\caption{Differentiable golden network for the Dyck-1 task, used in Experiment 3.}
\end{figure}

\begin{figure}[H]
\centering
\includegraphics[width=0.8\linewidth]{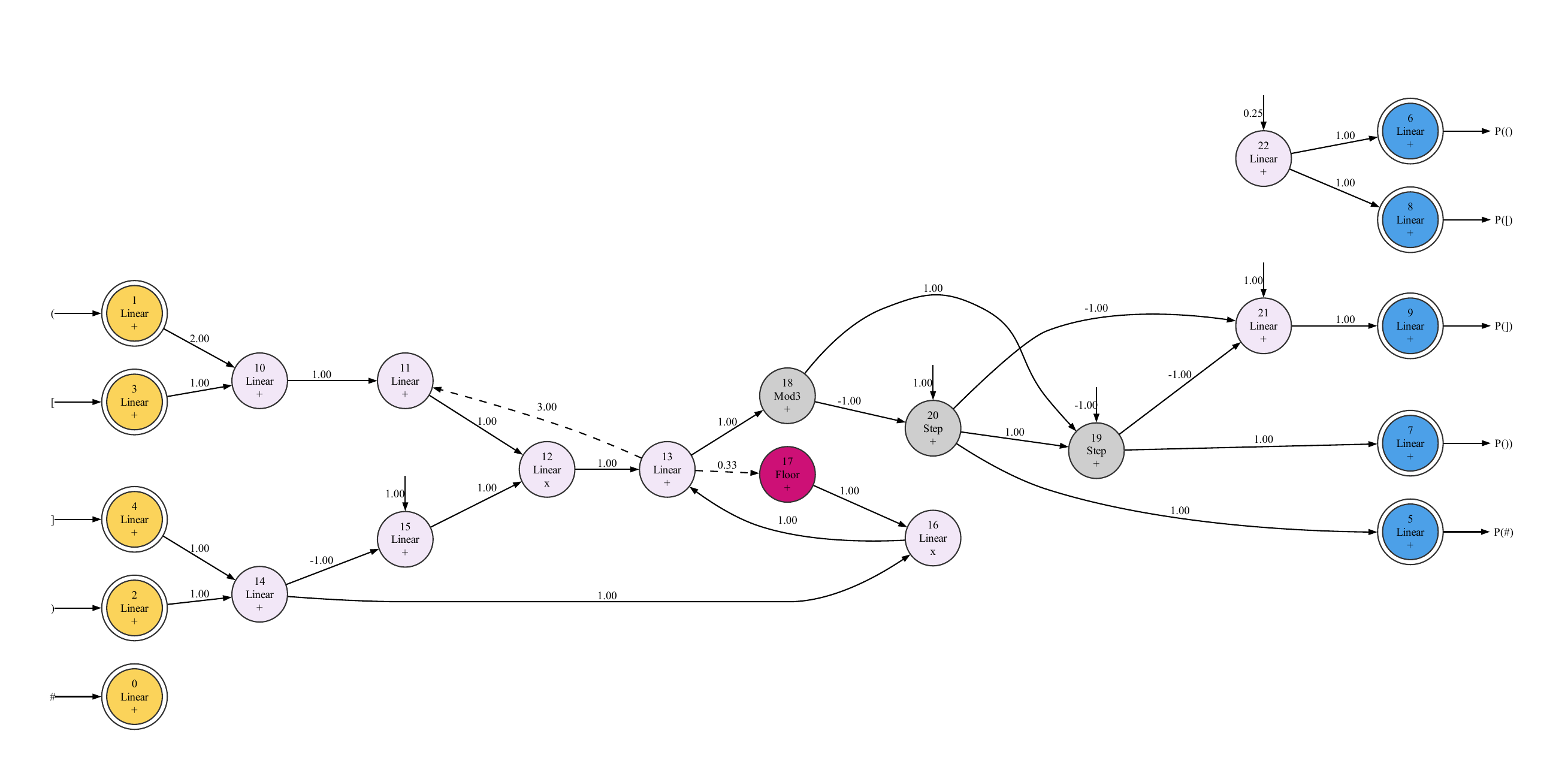}
\caption{Golden network for the Dyck-2 task.}
\end{figure}

\begin{figure}[H]
\centering
\includegraphics[width=0.8\linewidth]{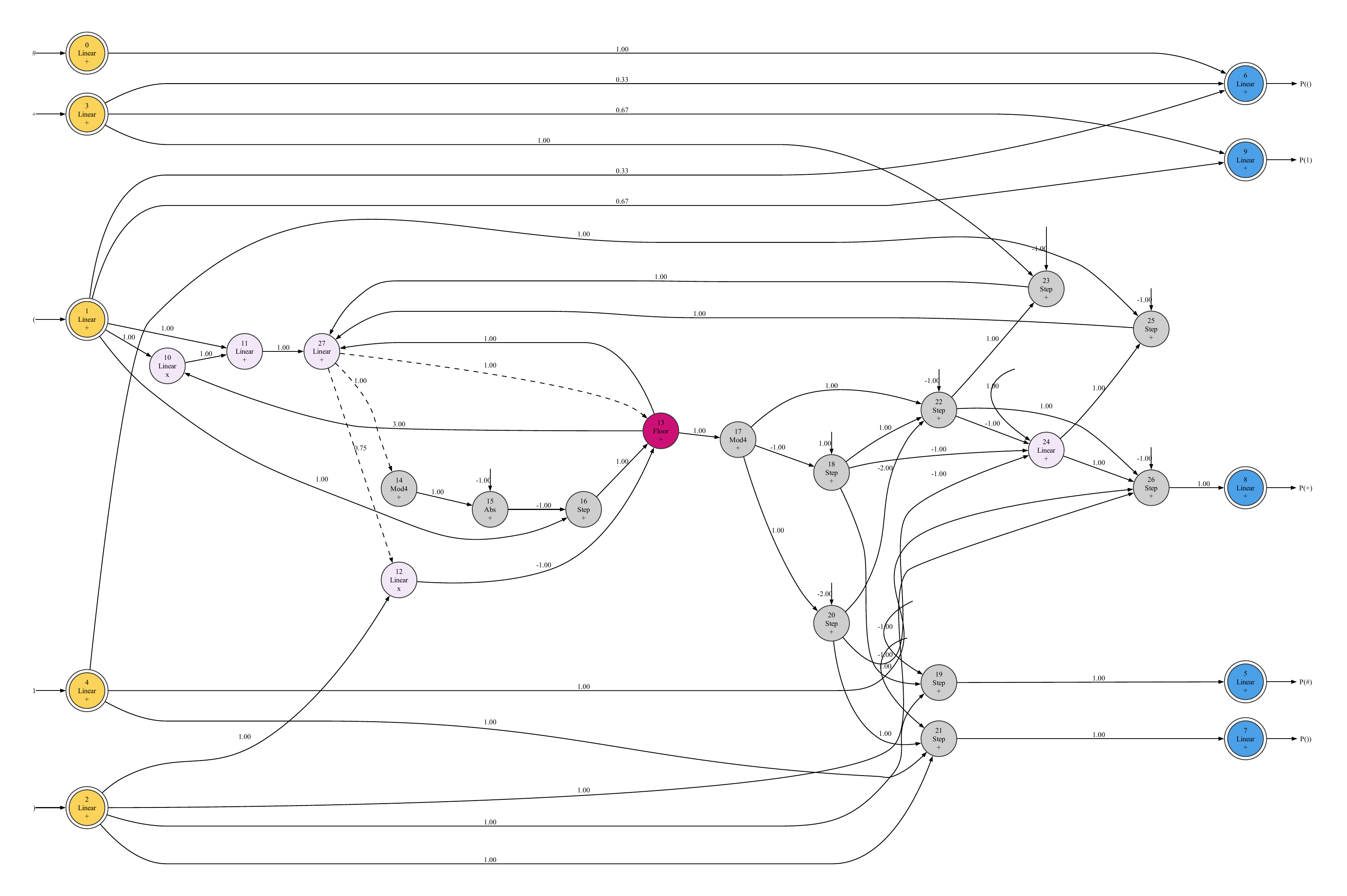}
\caption{Golden network for the Arithmetic task.}
\end{figure}

\begin{figure}[H]
\centering
\includegraphics[width=0.8\linewidth]{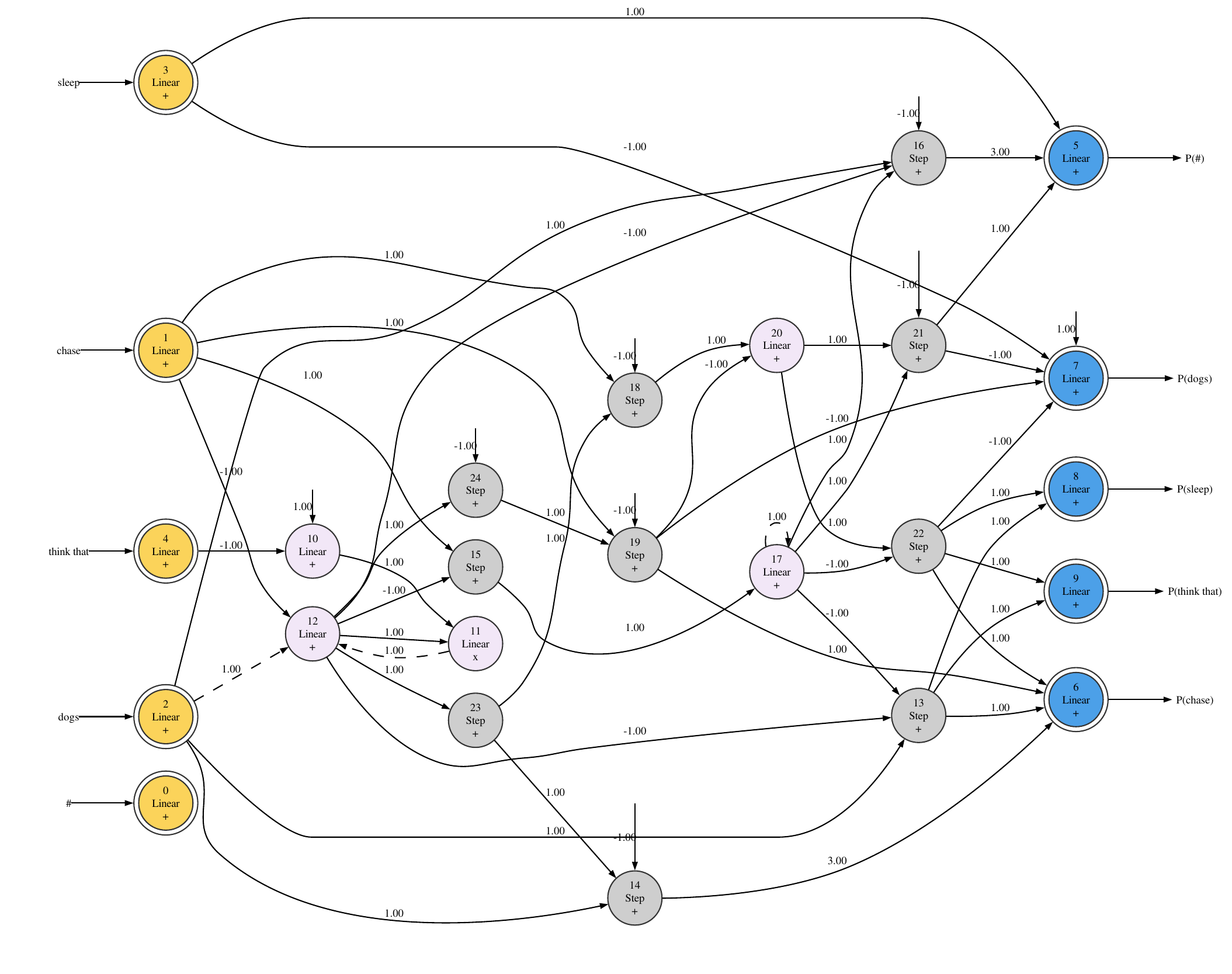}
\caption{Golden network for the Toy-English task.}
\end{figure}

\section{Genetic algorithm}
\label{app:gen-alg}

\begin{algorithm}[H]
	\caption{Genetic Algorithm for Evolving Networks}
	\label{gen-alg}
	\begin{algorithmic}[1]
		\Function{TournamentSelection}{$pop$, $objFunc$}
		\State $T \gets t$ random networks from $pop$
		\State $winner \gets \arg\min_{objFunc}(T)$
		\State $loser \gets \arg\max_{objFunc}(T)$
		\State \Return $winner, loser$
		\EndFunction
		
		\State $population \gets \emptyset$ \Comment{Population Initialization}
		\While{$|population| < N-1$}
		\State generate a random network $net$
		\State add $net$ to $population$
		\EndWhile
		\State add $goldenNet$ to $population$
		\Statex
		
		\State $generation \gets 0$ \Comment{Evolution Loop}
		\While{$generation < Gen$}
		\For{$N$ steps}
		\State $parent, loser \gets \textsc{TournamentSelection}(population, objFunc)$
		\State \textsc{eval}($offspring$) \Comment{Evaluate $offspring$ using $objFunc$}
		\State remove $loser$ from $population$
		\State add $offspring$ to $population$
		\EndFor
		\State $generation \gets generation + 1$
		\EndWhile
	\end{algorithmic}
\end{algorithm}

\section{Hyper-parameters}
\label{app:hyperparams}
\subsection{Genetic algorithm configuration}

\begin{table}[H]
\caption{Default genetic algorithm hyperparameters used in Experiments 1 and 2.}
\label{app:default-hparams}
\centering
\begin{adjustbox}{max width=\textwidth}
\begin{tabular}{ll}
\toprule
\textbf{Parameter} & \textbf{Value} \\
\midrule
Number of Islands & 500 \\
Training Batch Size & 500 \\
Migration ratio & 0.01 \\
Migration interval (seconds) & 1800 \\
Migration interval (generations) & 1000 \\
Number of generations & 25,000 \\
Population size & 500 \\
Elite ratio & 0.001 \\
Allowed activations & Linear, ReLU, Tanh \\
Allowed unit types & Summation \\
Tournament size & 2 \\
Mutation probability & 1.0 \\
Crossover probability & 0.0 \\
Maximum network units & 1024 \\
Experiment seed & 100 \\
Corpus seed & 100 \\
Golden network copies in initialization & 1 \\
\bottomrule
\end{tabular}
\end{adjustbox}
\end{table}

\subsection{Task-specific modifications}
Unless noted otherwise, all tasks inherit the default configuration from Table~\ref{app:default-hparams}. The grammars used for corpus generations are detailed in \ref{app:grammars}.

\begin{table}[H]
\caption{Overrides to allowed activation functions and unit types for each task.}
\centering
\begin{adjustbox}{max width=\textwidth}
\begin{tabular}{lll}
\toprule
\textbf{Task} & \textbf{Additional Allowed Activations} & \textbf{Additional Allowed Unit Types} \\
\midrule
$a^nb^n$ & Sigmoid, Unsigned Step & --- \\
$a^nb^nc^n$ & Sigmoid, Unsigned Step & --- \\
Dyck-1 & Sigmoid & --- \\
Dyck-2 & Floor, Modulo-3, Unsigned Step & Multiplication Unit \\
Arithmetic & Floor, Modulo-4, Absolute Value, Unsigned Step & Multiplication Unit \\
Toy-English & Unsigned Step & Multiplication Unit \\
\bottomrule
\end{tabular}
\end{adjustbox}
\end{table}

The maximum model complexity for simulations with limited $|H|$ is set to three times the complexity score $|H|$ of the corresponding golden network.

\subsection{Backpropagation configuration}
\label{app:backprop-hparams}

\begin{table}[H]
\caption{Backpropagation hyperparameters used in Experiment 3.}
\centering
\begin{adjustbox}{max width=\textwidth}
\begin{tabular}{ll}
\toprule
\textbf{Parameter} & \textbf{Value} \\
\midrule
Optimizer & Adam \\
Learning rate & $10^{-4}$ \\
Number of epochs & 1000 \\
Regularization methods & None, $L_1$, $L_2$ (with $\lambda = 1$) \\
Experiment seed & 100 \\
Corpus seed & 100 \\
\bottomrule
\end{tabular}
\end{adjustbox}
\end{table}

\section{Compute resources}
\label{app:compute-resource}
The genetic algorithm experiments were conducted on a SLURM cluster using CPU-only compute nodes. Each GA simulation was run on 252 CPU cores — 250 dedicated to GA islands and 2 reserved for logging and result collection. Runs were capped at 25,000 generations or 48 hours, whichever occurred first. 

The gradient descent experiments, being more lightweight, were executed on a personal computer and completed in under an hour.

\section{Additional figures}

\begin{figure}[H]
    \centering
    \includegraphics[width=0.8\linewidth]{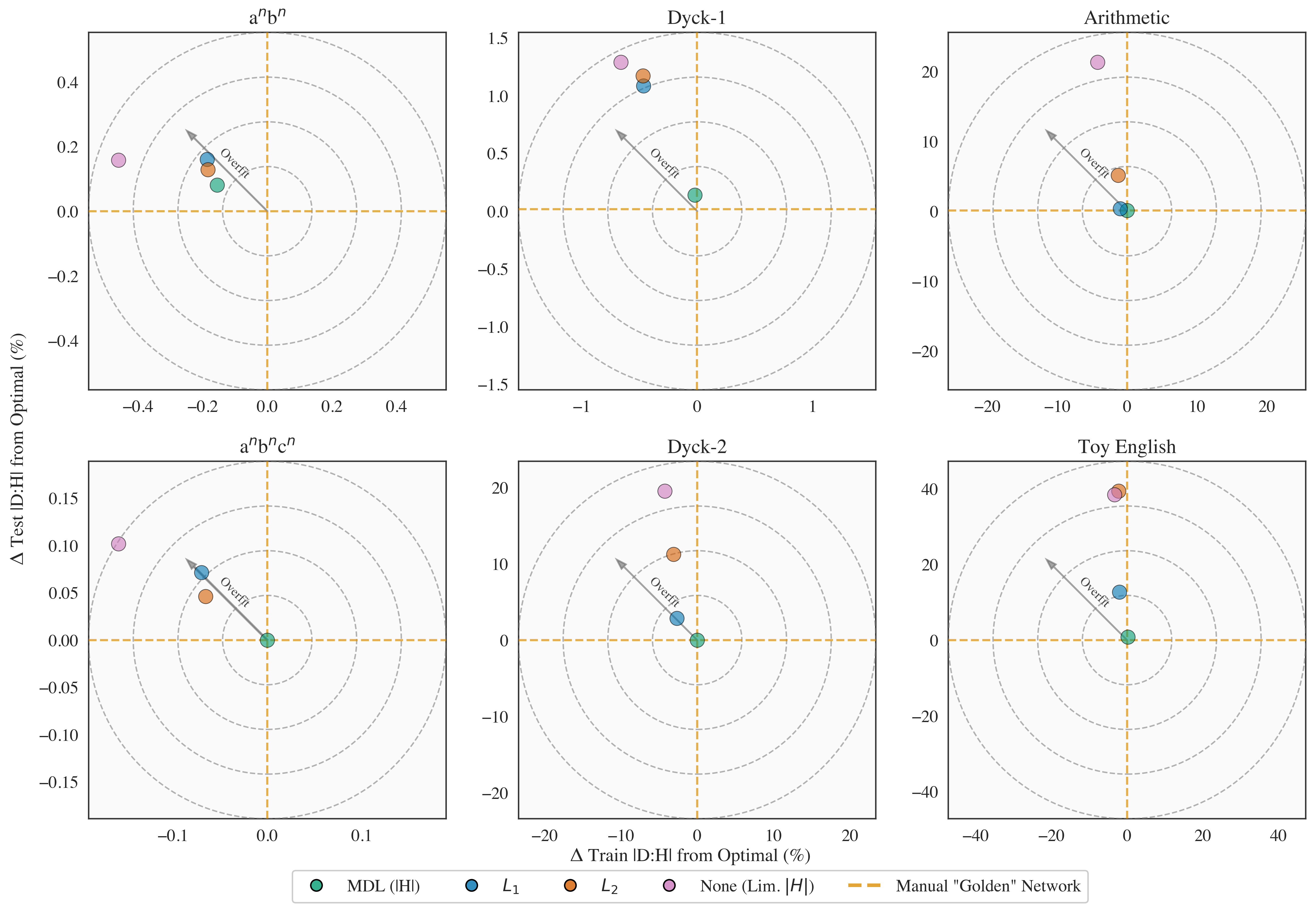}
    \caption{Relative deviation ($\Delta\%$) from optimal $|D:H|$ for each final network in Experiment 2, on train (x-axis) and test (y-axis), grouped by task. Proximity to center indicates better approximation of the analytical optimum.}
    \label{fig:weights-only-target-plot}
\end{figure}

\begin{figure}[H]
    \centering
    \includegraphics[width=0.9\linewidth]{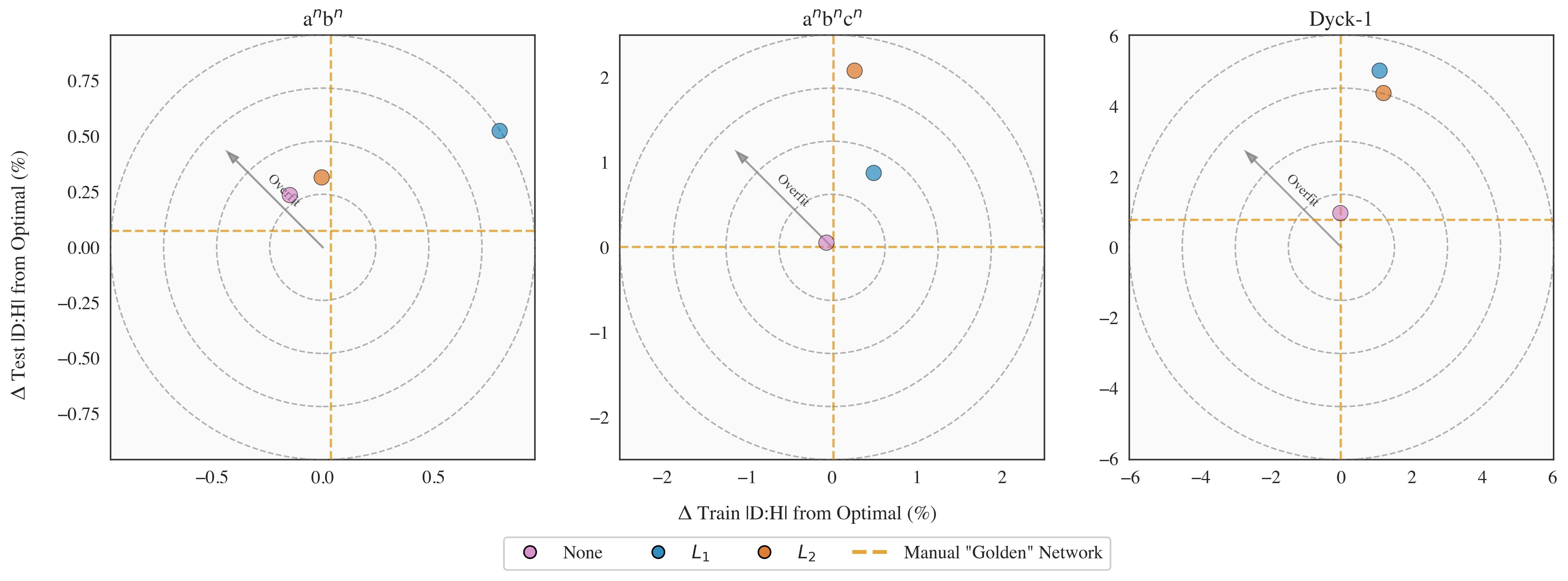}
    \caption{Relative deviation ($\Delta\%$) from optimal $|D:H|$ for each final network in Experiment 3, on train (x-axis) and test (y-axis), grouped by task. Proximity to center indicates better approximation of the analytical optimum.}
    \label{fig:gd-target-plot}
\end{figure}

\section{Full results}
\label{app:overlap}

\begin{sidewaystable}
    \caption{\textbf{GA architecture search full results.} For each network selected by GA, we report all regularization term values, irrespective of the regularizer used in optimization. Networks are evaluated on training samples, an exhaustive test set, and a non-overlapping test set (strings not seen during training), reporting the relative gap $\Delta (\%) = \frac{|D:H| - \text{Optimal}}{\text{Optimal}} \times 100$. * marks originally infinite scores due to zero‑probability errors. Simulations that were stopped due to the 48-hour time limit are marked with $\dagger$.}
    \label{tab:full-ga-arch-search-results}
	\centering
    \begin{adjustbox}{max width=\textheight}
\begin{tabular}{llccccccc}
\toprule
Task & Regularizer & $|H|$ & $L_1$ & $L_2$ & Train $|D:H|$ & Test $|D:H|$ & $\Delta^{Train}_{Optim}$ (\%) & $\Delta^{Test}_{Optim}$ (\%) \\
\midrule
a$^n$b$^n$ & (Golden) & (139) & (7.00) & (15.00) & (1531.77) & (2.94) & (0.0) & (0.0) \\
& MDL (|H|) & 139 & 7.00 & 15.00 & 1529.39 & 2.94 & \textbf{-0.2} & \textbf{0.1} \\
& $L_1$ & 460 & 5.46 & 10.54 & 1498.12 & 2.94$^*$ & -2.2 & 0.2 \\
& $L_2$ & 600 & 5.52 & 2.92 & 1514.66 & 3.27$^*$ & -1.1 & 11.4 \\
& None (Lim. $|H|$) & 430 & 33.95 & 110.80 & 1488.17 & 3.32$^*$ & -2.8 & 13.0 \\
\midrule
a$^n$b$^n$c$^n$ & (Golden) & (241) & (11.33) & (16.44) & (1483.91) & (2.94) & (0.0) & (0.0) \\
& MDL (|H|) & 241 & 12.33 & 19.44 & 1483.91 & 2.94 & \textbf{0.0} & \textbf{0.0} \\
& $L_1$ & 366 & 5.40 & 8.59 & 1482.88 & 2.94 & -0.1 & 0.1 \\
& $L_2$ & 602 & 5.49 & 3.25 & 1476.71 & 3.03$^*$ & -0.5 & 3.2 \\
& None (Lim. $|H|$) & 736 & 44.12 & 131.07 & 1460.67 & 2.96 & -1.6 & 0.6 \\
\midrule
Dyck-1 & (Golden) & (113) & (7.00) & (15.00) & (1544.48) & (1.77) & (-0.0) & (0.0) \\
& MDL (|H|) & 113 & 6.00 & 10.00 & 1544.22 & 1.77 & \textbf{-0.0} & \textbf{0.1} \\
& $L_1$ & 753 & 25.63 & 31.51 & 1488.17 & 1.90$^*$ & -3.6 & 7.5 \\
& $L_2$ & 1140 & 31.02 & 24.72 & 1479.37 & 1.96$^*$ & -4.2 & 10.7 \\
& None (Lim. $|H|$) & 353 & 38.43 & 189.06 & 1505.61 & 2.03$^*$ & -2.5 & 15.1 \\
\midrule
Dyck-2 & (Golden) & (579) & (27.33) & (35.11) & (2121.53) & (2.32) & (-0.0) & (0.0) \\
& MDL (|H|) & 327 & 15.00 & 23.50 & 2130.28 & 2.37 & \textbf{0.4} & \textbf{2.0} \\
& $L_1$ & 1629 & 35.95 & 36.82 & 2009.61 & 2.70$^*$ & -5.3 & 16.1 \\
& $L_2$ & 2248 & 42.04 & 36.04 & 1996.82 & 2.82$^*$ & -5.9 & 21.3 \\
& None (Lim. $|H|$)$^\dagger$ & 1757 & 94.99 & 204.82 & 1992.89 & 2.61$^*$ & -6.1 & 12.5 \\
\midrule
Arithmetic & (Golden) & (967) & (47.75) & (54.67) & (2581.29) & (3.96) & (0.0) & (0.1) \\
& MDL (|H|) & 431 & 20.25 & 25.31 & 2581.07 & 3.94 & \textbf{0.0} & \textbf{-0.4} \\
& $L_1$ & 1268 & 21.04 & 19.73 & 2540.22 & 3.99$^*$ & -1.6 & 1.0 \\
& $L_2$ & 1200 & 17.93 & 15.94 & 2560.62 & 4.01$^*$ & -0.8 & 1.4 \\
& None (Lim. $|H|$)$^\dagger$ & 3004 & 173.91 & 408.82 & 2477.42 & 4.23$^*$ & -4.0 & 7.0 \\
\midrule
Toy English & (Golden) & (870) & (49.00) & (61.00) & (2232.74) & (4.49) & (0.0) & (0.0) \\
& MDL (|H|) & 414 & 24.00 & 33.50 & 2231.96 & 4.57$^*$ & \textbf{-0.0} & \textbf{2.0} \\
& $L_1$ & 1215 & 15.06 & 14.18 & 2209.25 & 5.05$^*$ & -1.0 & 12.6 \\
& $L_2$ & 1330 & 13.15 & 8.23 & 2202.41 & 5.43$^*$ & -1.3 & 21.0 \\
& None (Lim. $|H|$) & 2625 & 163.22 & 390.45 & 2142.40 & 6.25$^*$ & -4.0 & 39.4 \\
\bottomrule
\end{tabular}
\end{adjustbox}
\end{sidewaystable}

\begin{sidewaystable}
    \caption{\textbf{GA weights only full results.} For each network selected by GA, we report all regularization term values, irrespective of the regularizer used in optimization. Networks are evaluated on training samples, an exhaustive test set, and a non-overlapping test set (strings not seen during training), reporting the relative gap $\Delta (\%) = \frac{|D:H| - \text{Optimal}}{\text{Optimal}} \times 100$. * marks originally infinite scores due to zero‑probability errors. Simulations that were stopped due to the 48-hour time limit are marked with $\dagger$.}
    \label{tab:full-weights-only-results}
	\centering
    \begin{adjustbox}{max width=\textheight}
\begin{tabular}{llccccccc}
\toprule
Task & Regularizer & $|H|$ & $L_1$ & $L_2$ & Train $|D:H|$ & Test $|D:H|$ & $\Delta^{Train}_{Optim}$ (\%) & $\Delta^{Test}_{Optim}$ (\%) \\
\midrule
a$^n$b$^n$ & (Golden) & (139) & (7.00) & (15.00) & (1531.77) & (2.94) & (0.0) & (0.0) \\
& MDL (|H|) & 139 & 7.00 & 15.00 & 1529.39 & 2.94 & \textbf{-0.2} & \textbf{0.1} \\
& $L_1$ & 217 & 3.96 & 5.43 & 1528.92 & 2.94 & -0.2 & 0.2 \\
& $L_2$ & 268 & 4.57 & 4.38 & 1528.96 & 2.94 & -0.2 & 0.1 \\
& None (Lim. $|H|$) & 385 & 95.18 & 2706.98 & 1524.71 & 2.94 & -0.5 & 0.2 \\
\midrule
a$^n$b$^n$c$^n$ & (Golden) & (241) & (11.33) & (16.44) & (1483.91) & (2.94) & (0.0) & (0.0) \\
& MDL (|H|) & 239 & 11.33 & 16.44 & 1483.91 & 2.94 & \textbf{0.0} & \textbf{0.0} \\
& $L_1$ & 362 & 5.07 & 8.36 & 1482.88 & 2.94 & -0.1 & 0.1 \\
& $L_2$ & 445 & 6.16 & 5.80 & 1482.94 & 2.94 & -0.1 & 0.0 \\
& None (Lim. $|H|$) & 733 & 758.90 & 246247.87 & 1481.57 & 2.94 & -0.2 & 0.1 \\
\midrule
Dyck-1 & (Golden) & (113) & (7.00) & (15.00) & (1544.48) & (1.77) & (-0.0) & (0.0) \\
& MDL (|H|) & 113 & 6.00 & 10.00 & 1544.22 & 1.77 & \textbf{-0.0} & \textbf{0.1} \\
& $L_1$ & 259 & 6.62 & 8.33 & 1537.35 & 1.79 & -0.5 & 1.1 \\
& $L_2$ & 408 & 7.66 & 5.27 & 1537.28 & 1.79 & -0.5 & 1.2 \\
& None (Lim. $|H|$) & 355 & 24.01 & 125.04 & 1534.32 & 1.79 & -0.7 & 1.3 \\
\midrule
Dyck-2 & (Golden) & (579) & (27.33) & (35.11) & (2121.53) & (2.32) & (-0.0) & (0.0) \\
& MDL (|H|) & 490 & 22.33 & 32.11 & 2121.53 & 2.32 & \textbf{-0.0} & \textbf{0.0} \\
& $L_1$ & 1202 & 31.29 & 33.04 & 2064.86 & 2.39$^*$ & -2.7 & 2.8 \\
& $L_2$ & 1426 & 32.54 & 29.38 & 2055.48 & 2.58$^*$ & -3.1 & 11.2 \\
& None (Lim. $|H|$) & 1731 & 124.94 & 1003.61 & 2031.26 & 2.78$^*$ & -4.3 & 19.5 \\
\midrule
Arithmetic & (Golden) & (967) & (47.75) & (54.67) & (2581.29) & (3.96) & (0.0) & (0.1) \\
& MDL (|H|) & 635 & 28.75 & 34.06 & 2581.29 & 3.96 & \textbf{0.0} & \textbf{0.1} \\
& $L_1^\dagger$ & 1420 & 16.88 & 15.22 & 2555.63 & 3.97$^*$ & -1.0 & 0.4 \\
& $L_2^\dagger$ & 1553 & 20.75 & 17.68 & 2548.16 & 4.15$^*$ & -1.3 & 5.1 \\
& None (Lim. $|H|$)$^\dagger$ & 2890 & 185.24 & 421.52 & 2471.51 & 4.79$^*$ & -4.2 & 21.3 \\
\midrule
Toy English & (Golden) & (870) & (49.00) & (61.00) & (2232.74) & (4.49) & (0.0) & (0.0) \\
& MDL (|H|) & 552 & 31.00 & 51.00 & 2237.59 & 4.52 & \textbf{0.2} & \textbf{0.8} \\
& $L_1$ & 1491 & 27.69 & 22.46 & 2186.61 & 5.05$^*$ & -2.1 & 12.7 \\
& $L_2$ & 1704 & 29.66 & 19.56 & 2182.59 & 6.25$^*$ & -2.2 & 39.3 \\
& None (Lim. $|H|$) & 2610 & 139.89 & 304.89 & 2158.71 & 6.21$^*$ & -3.3 & 38.4 \\
\bottomrule
\end{tabular}
\end{adjustbox}
\end{sidewaystable}

\begin{sidewaystable}
    \caption{\textbf{Gradient descent full results.} For each final network, we report all regularization term values, irrespective of the regularizer used in training. $|H|$ is approximated by the encoding length of the weights. Networks are evaluated on training samples, an exhaustive test set, and a non-overlapping test set (strings not seen during training), reporting the relative gap $\Delta (\%) = \frac{|D:H| - \text{Optimal}}{\text{Optimal}} \times 100$.}
    \label{tab:full-gd-results}
	\centering
    \begin{adjustbox}{max width=\textheight}
\begin{tabular}{llccccccc}
\toprule
Task & Regularizer & $\approx |H|$ & $L_1$ & $L_2$ & Train $|D:H|$ & Test $|D:H|$ & $\Delta^{Train}_{Optim}$ (\%) & $\Delta^{Test}_{Optim}$ (\%) \\
\midrule
a$^n$b$^n$ & (Golden) & (274) & (56.86) & (718.73) & (1532.33) & (2.94) & (0.0) & (0.1) \\
& None & 5524 & 57.87 & 728.90 & 1529.48 & 2.94 & -0.1 & 0.2 \\
& $L_1$ & 5854 & 56.17 & 707.85 & 1543.98 & 2.95 & 0.8 & 0.5 \\
& $L_2$ & 5610 & 56.30 & 707.86 & 1531.69 & 2.95 & -0.0 & 0.3 \\
\midrule
a$^n$b$^n$c$^n$ & (Golden) & (599) & (91.33) & (1655.78) & (1484.15) & (2.94) & (0.0) & (0.0) \\
& None & 20043 & 92.54 & 1661.71 & 1482.91 & 2.94 & -0.1 & 0.1 \\
& $L_1$ & 20541 & 90.06 & 1638.44 & 1491.15 & 2.96 & 0.5 & 0.9 \\
& $L_2$ & 19909 & 90.37 & 1638.25 & 1487.84 & 3.00 & 0.3 & 2.1 \\
\midrule
Dyck-1 & (Golden) & (148) & (20.75) & (104.56) & (1544.28) & (1.78) & (-0.0) & (0.8) \\
& None & 4876 & 21.14 & 107.30 & 1544.03 & 1.78 & -0.0 & 1.0 \\
& $L_1$ & 5138 & 20.06 & 100.91 & 1561.23 & 1.86 & 1.1 & 5.0 \\
& $L_2$ & 4874 & 20.20 & 100.91 & 1562.96 & 1.84 & 1.2 & 4.4 \\
\bottomrule
\end{tabular}
\end{adjustbox}
\end{sidewaystable}

\end{document}